# Identification of geometrical and elastostatic parameters of heavy industrial robots

A. Klimchik, Y. Wu, C. Dumas, S. Caro, B. Furet and A. Pashkevich

*Abstract—* The paper focuses on the stiffness modeling of heavy industrial robots with gravity compensators. The main attention is paid to the identification of geometrical and elastostatic parameters and calibration accuracy. To reduce impact of the measurement errors, the set of manipulator configurations for calibration experiments is optimized with respect to the proposed performance measure related to the end-effector position accuracy. Experimental results are presented that illustrate the advantages of the developed technique.

## I. INTRODUCTION

Further developments in aeronautic industry require high-precision and high-speed machining of huge aircraft components where industrial robots successfully replace conventional CNC-machines. For these applications, robots are quite attractive due to their large workspace (that can be easily extended) and high-speed motion capacity, as well as capability to process the parts with complex shape and geometry. However, because of high forces required for processing of modern aeronautic materials, the influence of robot stiffness becomes one of the key factors that essentially affect the position accuracy. To enhance robot stiffness, manufacturers tend to increase the manipulator link cross-sections that obviously leads to the augmentation of the robot mass. So, the gravity forces applied to the manipulator components become non-negligible and also contribute to the position errors. To overcome this difficulty, the robots are usually equipped with different types of gravity compensators, which, however, considerably complicate the stiffness modeling of those heavy manipulators.

The problem of stiffness modeling for the heavy manipulators with gravity compensators has been in the focus of rather limited number of works. In contrast, for conventional serial manipulators without gravity compensators, the problem has been studied by a number of authors that considered both industrial and medical robots with essential compliance in the links and joints. [1-3]. Relevant works are mainly based on the virtual joint method (VJM), which was firstly proposed by Salisbury [4] and lumped elastostatic properties of robot components in virtual springs. To our knowledge, the stiffness modeling for the manipulators with gravity compensators has not been studied in detail yet. Currently, the main activity in this area focuses on the gravity compensator design, which differs in kinematics [5] and/or may also employ some software tools embedded in the robot controller [6]. On the other hand, since the considered robots include closed loops induced by the compensators, some technique developed for the parallel manipulators can be adopted[7-9].

This paper proposes a VJM-based stiffness model for a serial manipulator with a compensator attached to the second joint. It is assumed that the gravity compensation torque is generated by a spring incorporated in an additional link, which creates the closed loop to be included in the stiffness model. The main attention is paid to the identification of the model parameters and calibration experiment planning. The developed approach is confirmed by the experimental results that deal with compliance error compensation for robotic cell employed in manufacturing of large dimensional aircraft components.

To address these problems the remainder of the paper is organized as follows. Section 2 presents the stiffness modeling background. In Section 3, the geometrical and elastostatic models of the gravity compensator are presented. Sections 4 and 5 are devoted to the geometrical and elastostatic calibration. Section 6 summarizes the main contributions.

## II. STIFFNESS MODELING BACKGROUND

Stiffness of a serial robot highly depends on its configuration and is defined by the Cartesian stiffness matrix that should be computed in the neighborhood of the loaded equilibrium configuration. Therefore, let us first present technique for computing the static equilibrium configuration and after focus on the computing of the stiffness matrix.

Using the VJM-based approach adopted in this paper, the manipulator can be presented as the sequence of rigid links separated by the actuators and virtual flexible joints incorporating all elastostatic properties of compliant elements [10]. Since for the considered robots weights of the link is not negligible, it should be also incorporated in the model. In the frame of the VJM-based technique, it is convenient to replace the link weights $\mathbf{P}_i$ by the equivalent pair of the forces applied to the both link ends $\mathbf{P}_i'$ and $\mathbf{P}_i''$. Further, after aggregation of all weight components applied to the same virtual joint, the external loading caused by the link weights can be described by the set of forces $\mathbf{G}_i$ concentrated in the virtual

*The work presented in this paper was partially funded by the Agence Nationale de la Recherche (ANR), France (Project ANR-2010-SEGI-003-02-COROUSSO).

A. Klimchik, A. Pashkevich and Y. Wu are with Ecole des Mines de Nantes, 4 rue Alfred-Kastler, Nantes 44307, France and with Institut de Recherches en Communications et Cybernétique de Nantes (IRCCyN), 44321 Nantes, France (phone: Tel.+33-251-85-83-00; fax.+33-251-85-83-49; e-mail: alexandr.klimchik@mines-nantes.fr, anatol.pashkevich@mines-nantes.fr, yier.wu@mines-nantes.fr).

S. Caro is with Centre National de la Recherche Scientifique (CNRS), France and with IRCCyN (e-mail: stephane.caro@irccyn.ec-nantes.fr).

C. Dumas, and B. Furet are with University of Nantes, France and with IRCCyN (e-mail: claire.dumas@univ-nantes.fr, benoit.furet@univ-nantes.fr).

joint centers. For computational convenience, they can be collected in the matrix $\mathbf{G} = [\mathbf{G}_1 ... \mathbf{G}_n]$ that will be referred to as "the auxiliary loading". This allows us to distinguish impact of the gravity from the influence of the conventional loading $\mathbf{F}$ applied to the robot end-effector (that can be caused by the cutting forces in milling, for instance). Using those notations and the above defined assumptions the elastostatic model of the heavy serial robot can be presented as shown in Fig. 1.

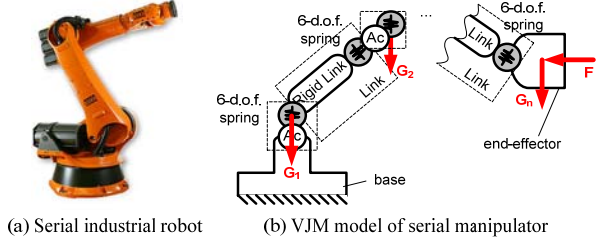

(a) Serial industrial robot  (b) VJM model of serial manipulator

Figure 1. Virtual joint model of a serial industrial robot

For this serial chain, the end-effector location $\mathbf{t}$ and locations of the node centers $\mathbf{t}_j$ can be defined by the vector functions $\mathbf{g}(\mathbf{q}, \boldsymbol{\theta})$ and $\mathbf{g}_j(\mathbf{q}, \boldsymbol{\theta})$, where $\mathbf{q}$, $\boldsymbol{\theta}$ denote the vectors of the actuator and virtual joint coordinates respectively. It can be proved that applying the principle of virtual work and linearizing the geometrical model, the static equilibrium equations can be written in a matrix form as [11]

$$\mathbf{J}_\theta^{(G)T} \cdot \mathbf{G} + \mathbf{J}_\theta^{(F)T} \cdot \mathbf{F} = \mathbf{K}_\theta \cdot \boldsymbol{\theta} \qquad (1)$$

where $\mathbf{J}_\theta^{(G)} = \left[\mathbf{J}_\theta^{(1)T} ... \mathbf{J}_\theta^{(n)T}\right]^T$ and $\mathbf{J}_\theta^{(F)} = \mathbf{J}_\theta^{(n)}$ denote the Jacobians computed with respect to the relevant nodes where the forces are applied to, i.e. $\mathbf{J}_\theta^{(j)} = \partial \mathbf{g}_j(\mathbf{q}, \boldsymbol{\theta})/\partial \boldsymbol{\theta}$; and the matrix $\mathbf{K}_\theta = diag\left(\mathbf{K}_{\theta_1}, ..., \mathbf{K}_{\theta_n}\right)$ aggregates stiffnesses of all virtual springs.

To find the desired static equilibrium configuration, the above system should be solved subject to the geometrical constraint $\mathbf{t} = \mathbf{g}(\mathbf{q}, \boldsymbol{\theta})$, where the end-effector location is assumed to be given and the force $\mathbf{F}$ is treated as an unknown. It is obvious that it is a dual problem compared to the traditional one where $\mathbf{F}$ is given and the vector $\mathbf{t}$ is unknown, but this approach essentially reduces the computational effort. Relative iterative algorithm for solving this system can be presented as

$$\mathbf{F}_{i+1} = \left(\mathbf{J}_\theta^{(F)} \cdot \mathbf{K}_\theta^{-1} \cdot \mathbf{J}_\theta^{(F)T}\right)^{-1} \cdot$$
$$\left(\mathbf{t}_{i+1} - \mathbf{g}(\mathbf{q}, \boldsymbol{\theta}_i) + \mathbf{J}_\theta^{(F)} \boldsymbol{\theta}_i - \mathbf{J}_\theta^{(F)} \mathbf{K}_\theta^{-1} \mathbf{J}_\theta^{(G)T} \mathbf{G}_i\right) \qquad (2)$$
$$\boldsymbol{\theta}_{i+1} = \mathbf{K}_\theta^{-1} \left(\mathbf{J}_\theta^{(G)T} \cdot \mathbf{G}_i + \mathbf{J}_\theta^{(F)T} \cdot \mathbf{F}_{i+1}\right)$$

This algorithm allows us to find the vectors $\mathbf{F}$, $\boldsymbol{\theta}$ defining deflections in the virtual springs under the loading $\mathbf{F}$, which will be required for computing the stiffness matrix.

To find the corresponding stiffness matrix $\mathbf{K}_C$, the force-deflection relation (1) should be linearized in the neighborhood of the equilibrium configuration $(\mathbf{F}, \boldsymbol{\theta}, \mathbf{t})$. After relevant transformations [11], one can get the following expression

$$\mathbf{K}_C = \left(\mathbf{J}_\theta^{(F)}(\mathbf{K}_\theta - \mathbf{H}_{\theta\theta})^{-1}\mathbf{J}_\theta^{(F)T}\right)^{-1} \qquad (3)$$

which includes both the first and second order derivatives (Jacobians and Hessians) of the functions $\mathbf{g}_j(\mathbf{q}, \boldsymbol{\theta})$ describing the manipulator geometry: $\mathbf{J}_\theta^{(F)} = \partial \mathbf{g}(\mathbf{q}, \boldsymbol{\theta})/\partial \boldsymbol{\theta}$, $\mathbf{H}_{\theta\theta} = \mathbf{H}_{\theta\theta}^{(F)} + \mathbf{H}_{\theta\theta}^{(G)} + \mathbf{J}_\theta^{(G)T} \partial \mathbf{G}/\partial \boldsymbol{\theta}$, $\mathbf{H}_{\theta\theta}^{(G)} = \sum_{j=1}^{n} \partial^2 \mathbf{g}_j^T \mathbf{G}_j / \partial \boldsymbol{\theta}^2$, $\mathbf{H}_{\theta\theta}^{(F)} = \partial^2 \mathbf{g}^T \mathbf{F} / \partial \boldsymbol{\theta}^2$.

It should be noted that the above presented expressions have been derived for the case of manipulators without gravity compensators, where the matrix $\mathbf{K}_\theta$ is constant. In the following Section, to take into account the compensator influence, these expressions will be modified by using the configuration dependent joint stiffness matrix $\mathbf{K}_\theta(\mathbf{q})$ describing properties of the virtual springs.

### III. MECHANICS OF GRAVITY COMPENSATOR

The mechanical structure of the gravity compensator under study is presented in Fig. 2. The compensator incorporates a passive spring attached to the first and second links, which creates a closed loop that generates the torque applied to the second joint of the manipulator. This design allows us to limit the stiffness model modification by incorporating in it the compensator torque $M_c$ and adjusting the virtual joint stiffness matrix $\mathbf{K}_\theta$ that here depends on the second joint variable $q_2$ only.

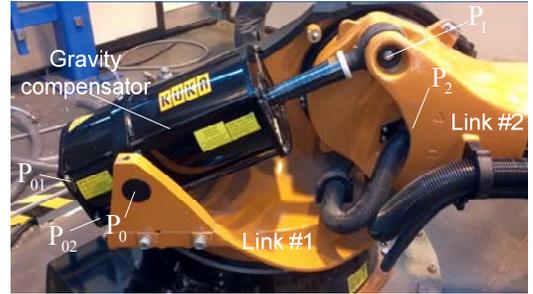

(a) gravity compensator of robot KUKA KR-270 TM

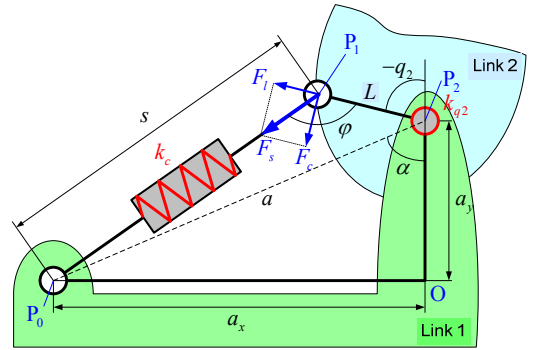

(b) model of gravity compensator

Figure 2. Gravity compensator and its model

The compensator geometrical model includes three node points $P_0$, $P_1$, $P_2$, where two distances $|P_1, P_2|$, $|P_0, P_2|$ are constants and the third one $|P_0, P_1|$ varies and depends on $q_2$. Let us denote them $L = |P_1, P_2|$, $a = |P_1, P_2|$, $s = |P_1, P_2|$. Besides, let us introduce the angles $\alpha$, $\varphi$ and the distances $a_x$ and $a_y$, whose geometrical meaning is clear from Fig. 2. Using these notations, the variable $s$ describing the compensator spring deflection can be computed from the expression

$$s^2 = a^2 + L^2 + 2 \cdot a \cdot L \cdot \cos(\alpha - q_2) \qquad (4)$$

which defines the function $s(q_2)$.

This mechanical design allows to balance the manipulator weight for any given configuration by adjusting the compensator spring preloading. It can be taken into account by introducing the zero-value of the compensator length $s_0$ corresponding to the unloaded spring. Under this assumption, the compensator force applied to the node $P_1$ can be expressed as follows

$$F_s = K_c \cdot (s - s_0) \qquad (5)$$

where $k_c$ is the compensator spring compliance.

Further, the angle $\varphi$ between the compensator links $P_0P_1$ and $P_1P_2$ (see Fig. 2) can be found from the expression

$$\sin\varphi = a/s \cdot \sin(\alpha - q_2) \qquad (6)$$

which allows us to compute the compensator torque $M_c$ applied to the second joint

$$M_c = K_c \cdot (1 - s_0/s) \cdot a \cdot L \cdot \sin(\alpha - q_2) \qquad (7)$$

Upon differentiation of the latter expression with respect to $q_2$, the equivalent stiffness of the second joint (comprising both the manipulator and compensator stiffnesses) can be expressed as:

$$K_{\theta_2} = K_{\theta_2}^0 + K_c \cdot a L \cdot \eta_{q_2} \qquad (8)$$

where the coefficient

$$\eta_{q_2} = \frac{s_0}{s}\left(\frac{aL}{s^2}\sin^2(\alpha - q_2) + \cos(\alpha - q_2)\right) - \cos(\alpha - q_2) \qquad (9)$$

highly depends on the value of joint variable $q_2$ and the initial preloading in the compensator spring described by $s_0$. To illustrate this property, Fig. 3 presents a set of curves $\eta(q_2)$ obtained for different values of $s_0$ (the remaining parameters $\alpha$, $a$, $L$ correspond to robot KUKA KR-270 studied in the experimental part of the paper, see Section IV).

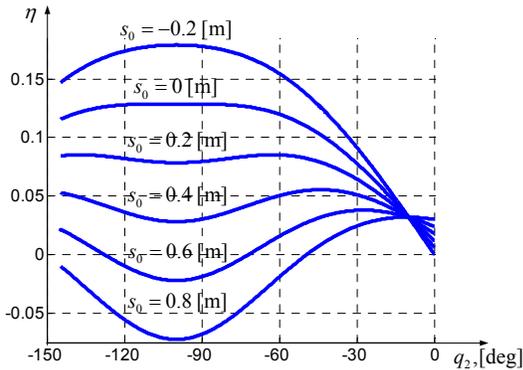

Figure 3. Variation of the gravity compensator impact on the equivalent stiffness of the second joint

Hence, using expression (8), it is possible to extend the classical stiffness model of the serial manipulator presented in Section II by modifying the virtual spring parameters in accordance with the compensator properties. While in the paper this approach has been used for the particular compensator type (spring-based, acting on the second joint), the similar idea can be evidently applied to other compensator types.

Summarizing this Section, it is worth mentioning that the geometrical and elastostatic models of a heavy manipulator with a gravity compensator should include some additional parameters ($\alpha$, $a$, $L$ and $K_c$, $s_0$ for the presented case) that are usually not included in datasheets. For this reason, the following Sections focus on the identification of the extended set of manipulator parameters.

## IV. IDENTIFICATION OF COMPENSATOR GEOMETRY

In contrast to the serial manipulator that can be treated as a principal mechanism of the considered robots (whose geometry is usually defined in datasheets and can be perfectly tuned by means of calibration [12][13]), geometrical data concerning gravity compensators are usually not included in the technical documentation provided by the robot manufacturers. For this reason, this Section focuses on the identification of the geometrical parameters for the described above compensator mechanism (see Fig. 2).

### A. Methodology

The geometrical structure of the considered gravity compensator is presented in Fig. 4. Its principal geometrical parameters are denoted as $L$, $a_x$, $a_y$, where $a_x = a \cdot \cos\alpha$, $a_y = a \cdot \sin\alpha$ (see notation in Section III). As follows from the figure, the identification problem can be reduced to the determination of relative locations of points $P_0$ and $P_1$ with respect to $P_2$.

It is assumed that the measurement data are provided by the laser tracker whose "world" coordinate system is located at the intersection of the first and second actuated manipulator joints. The axes Y, Z of this system are aligned with the axes of joints #1 and #2 respectively, while the axis X is directed to ensure right-handed orthogonal basis. To obtain required data, there are several markers attached to the compensator mechanism (see Fig. 4). The first one is located at point $P_1$, which is easily accessible and perfectly visible (the center of the compensator axis $P_1$ is exactly ticked on the fixing element). In contrast, for the point $P_0$, it is not possible to locate the marker precisely. For this reason, several markers $P_{0i}$ are used that are shifted with respect to $P_0$, but located on the rigid component of the compensator mechanism (these markers are rotating around $P_0$ while the joint coordinate $q_2$ is actuated). It should be noted that for the adopted compensator geometrical model (which is in fact a planar one), the marker location relative to the plane XY is not significant, since the identification algorithm presented in the following sub-section will ignore Z-coordinate.

Using this setting, the identification problem is solved in two steps. The first step is devoted to the identification of the relative location of points $P_1$ and $P_2$. Here, for different values of the manipulator joint coordinates $\{q_{2i}, i = \overline{1,m}\}$, the laser tracker provides the set of the vectors $\{\mathbf{p}_1^i\}$ describing the points that are located in an arc of the circle. After matching these points with a circle, one can obtain the desired value of $L$ (circle radius) and the Cartesian coordinates $\mathbf{p}_2$ of the point $P_2$ (circle center) with respect to the laser tracker coordinate system.

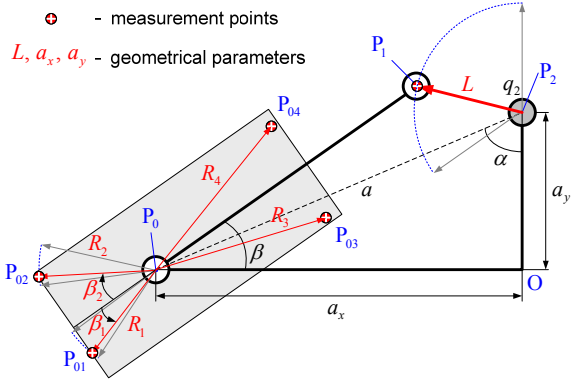

Figure 4. Geometrical parameters of the gravity compensator and location of the measurement points labeled with markers

The second step deals with the identification of the relative location of points $P_0$ and $P_2$. Relevant information is extracted from two data sets $\{\mathbf{p}_{01}^i\}$ and $\{\mathbf{p}_{02}^i\}$ that are provided by the laser tracker while targeting at the markers $P_{01}$ and $P_{02}$. Here, the points are matched to two circle arcs with the same center (explicitly assuming that the compensator model is planar), which yields the Cartesian coordinates $\mathbf{p}_0$ of the point $P_0$, also with respect to the laser tracker coordinate system. Finally, the desired values $a_x$, $a_y$ are computed as a projection of the difference vector $\mathbf{a} = \mathbf{p}_2 - \mathbf{p}_0$ on the corresponding axis of the coordinate system.

As follows from the presented methodology, a key numerical problem in the presented approach is the matching of the experimental points with a circle arc. It looks like a classical problem, however, there is a particularity here caused by availability of additional data $\{q_{2i}\}$ describing relative locations of the points $\{\mathbf{p}_1^i\}$. This feature allows us to reformulate the identification problem and to achieve higher accuracy compared with the traditional approach.

### B. Identification algorithm

The above presented methodology requires solution of two identification problems. The first one aims at approximating of a given set of points (with additional arc angle argument) with an arc circle, which provides the circle center and the circle radius. The second problem deals with an approximation of several sets of points by corresponding number of circle arcs with the same center. Let us consider them sequentially.

To match the given set of points $\{\mathbf{p}_i\}$ with additional set of angles $\{q_i\}$ with a circle arc, let us define the affine mapping

$$\mathbf{p}_i = \mu \mathbf{R} \mathbf{u}_i + \mathbf{t} \qquad (10)$$

where $\mathbf{u}_i = [\cos q_i, \sin q_i, 0]^T$ denotes the set of reference points located on the unit circle whose distribution on the arc is similar to $\mathbf{p}_i$, $\mu$ is the scaling factor that defines the desired circle radius, $\mathbf{R}$ is the orthogonal rotation matrix, $\mathbf{t}$ is the vector of the translation that defines the circle center. It worth mentioning that such a formulation has an advantage (in the sense of accuracy) comparing to a traditional circle approximation and it is a generalization of Procrustes problem known from the matrix analysis.

Using equation (10), the identification can be reduced to the following optimization problem

$$F = \sum_{i=1}^m (\mathbf{p}_i - \mu \mathbf{R} \mathbf{u}_i - \mathbf{t})^T (\mathbf{p}_i - \mu \mathbf{R} \mathbf{u}_i - \mathbf{t}) \to \min_{\mu, \mathbf{R}, \mathbf{t}} \qquad (11)$$

which should be solved subject to the orthogonality constraint $\mathbf{R}^T \mathbf{R} = \mathbf{I}$. After differentiation with respect to $\mathbf{t}$, the latter variable can be expressed as

$$\mathbf{t} = m^{-1} \sum_{i=1}^m \mathbf{p}_i - \mu m^{-1} \mathbf{R} \sum_{i=1}^m \mathbf{u}_i \qquad (12)$$

That leads to the simplification of (11) to

$$F = \sum_{i=1}^m (\hat{\mathbf{p}}_i - \mu \mathbf{R} \hat{\mathbf{u}}_i)^T (\hat{\mathbf{p}}_i - \mu \mathbf{R} \hat{\mathbf{u}}_i) \to \min_{r, \mathbf{R}} \qquad (13)$$

where

$$\hat{\mathbf{p}}_i = \mathbf{p}_i - m^{-1} \sum_{i=1}^m \mathbf{p}_i; \qquad \hat{\mathbf{u}}_i = \mathbf{u}_i - m^{-1} \sum_{i=1}^m \mathbf{u}_i \qquad (14)$$

Further, differentiation with respect to $\mu$ yields to

$$\mu = \sum_{i=1}^m \hat{\mathbf{p}}_i^T \mathbf{R} \hat{\mathbf{u}}_i \Big/ \sum_{i=1}^m \hat{\mathbf{u}}_i^T \hat{\mathbf{u}}_i \qquad (15)$$

So, finally, after relevant substitutions the objective function can be presented as

$$F = \sum_{i=1}^m \hat{\mathbf{p}}_i^T \hat{\mathbf{p}}_i - \left( \sum_{i=1}^m \hat{\mathbf{u}}_i^T \hat{\mathbf{u}}_i \right)^{-1} \left( \sum_{i=1}^m \hat{\mathbf{p}}_i^T \mathbf{R} \hat{\mathbf{u}}_i \right)^2 \to \min_{\mathbf{R}} \qquad (16)$$

where the unknown matrix $\mathbf{R}$ must satisfy the orthogonality constraint $\mathbf{R}^T \mathbf{R} = \mathbf{I}$. Since the matrix $\mathbf{R}$ is included in the second term only, the problem can be further simplified to

$$F' = \sum_{i=1}^m \hat{\mathbf{p}}_i^T \mathbf{R} \hat{\mathbf{u}}_i = trace\left( \mathbf{R} \sum_{i=1}^m \hat{\mathbf{u}}_i \hat{\mathbf{p}}_i^T \right) \to \max_{\mathbf{R}} \qquad (17)$$

and can be solved using SVD-decomposition of the matrix

$$\sum_{i=1}^m \hat{\mathbf{u}}_i \hat{\mathbf{p}}_i^T = \mathbf{U} \mathbf{\Sigma} \mathbf{V}^T \qquad (18)$$

where the matrices $\mathbf{U}, \mathbf{V}$ are orthogonal and $\mathbf{\Sigma}$ is the diagonal matrix of the singular values. Further, using the same approach as for the Procrustes problem, it can be proved that the desired rotation matrix can be computed as

$$\mathbf{R} = \mathbf{V} \mathbf{U}^T \qquad (19)$$

which sequentially allows to find the scaling factor $\mu$ defining the arc radius and the vector $\mathbf{t}$ defining the arc center.

The second problem aims at approximating of several point sets $\{\mathbf{p}_i^1\}, ..., \{\mathbf{p}_i^k\}$ by corresponding number of concentric circle arcs with the same center $\mathbf{p}_0$. It should be noted that here the data set $\{q_i\}$ is not useful, since the required angles $\{\beta_i\}$ are not measured directly and cannot be computed without having exact compensator geometry. In this case, the objective function can be written in a straightforward way

$$F = \sum_{j=1}^k \sum_{i=1}^m \left( R_j^2 - (\mathbf{p}_i^j - \mathbf{p}_0)^T (\mathbf{p}_i^j - \mathbf{p}_0) \right)^2 \to \min_{\mathbf{p}_0, R_j} \qquad (20)$$

But it can be proved that this optimization problem does not lead to a unique solution (in fact, it gives the rotation axis passing through the desired center). For this problem, differentiation of the objective function $F$ with respect to $R_j^2$ yields

$$R_j^2 = m^{-1} \sum_{i=1}^{m} \left(\mathbf{p}_i^j - \mathbf{p}_0\right)^T \left(\mathbf{p}_i^j - \mathbf{p}_0\right) \quad (21)$$

which after substitution into (20) allows us to rewrite the problem in the following way

$$F = \sum_{j=1}^{k} \sum_{i=1}^{m} \left(2\mathbf{p}_0^T \hat{\mathbf{p}}_i^j - \hat{s}_i^j\right)^2 \to \min_{\mathbf{p}_0} \quad (22)$$

where

$$\hat{\mathbf{p}}_i^j = \mathbf{p}_i^j - m^{-1} \sum_{l=1}^{m} \mathbf{p}_l^j; \quad \hat{s}_i^j = \mathbf{p}_i^{jT} \mathbf{p}_i^j - m^{-1} \sum_{l=1}^{m} \mathbf{p}_l^{jT} \mathbf{p}_l^j \quad (23)$$

Further, after differentiation with respect to $\mathbf{p}_0$, one can get the underdetermined system of linear equations

$$\left(\sum_{j=1}^{k} \sum_{i=1}^{m} \hat{\mathbf{p}}_i^j \hat{\mathbf{p}}_i^{jT}\right) \mathbf{p}_0 = \frac{1}{2} \sum_{j=1}^{k} \sum_{i=1}^{m} \hat{s}_i^j \hat{\mathbf{p}}_i^j \quad (24)$$

whose solution

$$\mathbf{p}_0 = \mathbf{p}_c + \xi \mathbf{n} \quad (25)$$

is expressed via the rotation axis vector $\mathbf{n}$ and a point belonging to this axis $\mathbf{p}_c$ (here $\xi$ is an arbitrary scalar factor).

To solve this ambiguity, an additional objective should be defined

$$\sum_{j=1}^{k} R_j^2 \to \min_{R_j} \quad (26)$$

which leads to the following solution for the scalar parameter

$$\xi = \mathbf{n}^T \left(-\mathbf{p}_c + k^{-1} m^{-1} \sum_{j=1}^{k} \sum_{i=1}^{m} \mathbf{p}_i^j\right) \quad (27)$$

and for the vector

$$\mathbf{p}_c = \frac{1}{2} \left(\sum_{j=1}^{k} \sum_{i=1}^{m} \hat{\mathbf{p}}_i^j \hat{\mathbf{p}}_i^{jT}\right)^{-1} \sum_{j=1}^{k} \sum_{i=1}^{m} \hat{s}_i^j \hat{\mathbf{p}}_i^j \quad (28)$$

So, the desired arc center is expressed as follows

$$\mathbf{p}_0 = \left(\mathbf{I} - \mathbf{n}\mathbf{n}^T\right) \mathbf{p}_c + k^{-1} m^{-1} \mathbf{n}\mathbf{n}^T \sum_{j=1}^{k} \sum_{i=1}^{m} \mathbf{p}_i^j \quad (29)$$

It should be mentioned that practical application of the latter expression is essentially simplified by the adopted assumption concerning orientation of the reference coordinate system (see previous sub-section), the direction of the identified rotation axis is close to Z-direction.

Hence, the developed algorithms allows us to identify the compensator geometrical parameters $L$, $a_x$, $a_y$ that are directly related to the above mentioned rotation center points $P_0$, $P_2$ and corresponding radii. Below they will be applied to the processing of the experimental data.

## C. Experimental results

To demonstrate efficiency of the developed technique, the experimental study has been carried out. The experimental setup employed the robot KR-270 and the Leica laser tracker, which allowed us to measure the Cartesian coordinates of the markers attached to the compensator elements (see Figs 2, 4). Six different manipulator configurations where considered that differed in the value of the joint angle $q_2$ and three markers has been used. The experimental data are presented in Table I.

These data has been processed using the identification algorithm presenting in the previous sub-section. The obtained values for the parameters of interest $L$, $a_x$, $a_y$ are given in Table II. It also includes the confidence intervals computed as $\pm 3\sigma$, where the standard deviation $\sigma$ has been evaluated using the Gibbs sampling. In the next section, the obtained model will be used for some transformations required during elastostatic calibration

TABLE I. EXPERIMENTAL DATA FOR GEOMETRICAL CALIBRATION

| $q_2$ [deg] | $P_1$ | | $P_{01}$ | | $P_{02}$ | |
|---|---|---|---|---|---|---|
| | x, [mm] | y, [mm] | x, [mm] | y, [mm] | x, [mm] | y, [mm] |
| -0.01 | -31.84 | 183.86 | -872.10 | -125.38 | -813.50 | -255.59 |
| -30 | -118.44 | 143.42 | -872.30 | -126.07 | -813.33 | -256.18 |
| -60 | -173.30 | 65.12 | -872.50 | -109.90 | -825.09 | -244.64 |
| -90 | -181.76 | -30.14 | -868.43 | -78.20 | -844.66 | -219.04 |
| -120 | -141.45 | -116.82 | -858.90 | -47.60 | -859.43 | -190.44 |
| -145 | -78.10 | -165.47 | -852.53 | -33.68 | -864.66 | -176.01 |

TABLE II. GEOMETRICAL PARAMETERS OF GRAVITY COMPENSATOR

| | L, [mm] | $a_x$, [mm] | $a_y$, [mm] |
|---|---|---|---|
| value | 184.72 | 685.93 | 120.30 |
| CI | ±0.06 | ±0.70 | ±0.69 |

## V. ELASTOSTATIC PARAMETERS IDENTIFICATION

In contrast to geometrical calibration, where the manipulator and compensator can be considered independently, in elastostatic calibration the corresponding equations cannot be separated and the model parameters should be identified simultaneously. This section gives general ideas of the developed methodology and relevant identification algorithms, and also presents experimental results validating the proposed technique.

### A. Methodology

In the frame of the adopted VJM-based modeling approach the desired stiffness parameters describe elasticity of the virtual springs located in the actuated joints of the manipulator, and also elasticity and preloading of the compensator spring (see Section III for details). Let us denote them as $k_{\theta_j}$, $j = 1, 6$ for the manipulator joint compliances and $k_c$, $s_0$ for the compliance and preloading of the compensator.

To find the desired parameters, the manipulator sequentially passes through several measurement configurations where the external loading is applied to the special end-effector described in Fig. 5 (it allows to generate both forces and torques applied to the manipulator). Using the laser tracker, the Cartesian coordinates of the reference points are measured twice, before and after loading. To increase identification accuracy, it is preferable to use several reference points (markers) and to apply the loading of the maximum allowed magnitude. It is worth mentioning that in order to avoid numerical singularities, the direction of the external loading should not be the same for all experiments (in spite of the fact that the gravity-based loading is the most attractive from the practical point of view). Thus, the calibration experiments yield the dataset that includes values of the manipulator joint coordinates $\{\mathbf{q}_i\}$, applied forces/torques $\{\mathbf{F}_i\}$ and corresponding deflections of the reference points $\{\Delta\mathbf{p}_i\}$. Using these data, the elastostatic parameters of $k_{\theta_j}$, $j=1,6$ and $k_c$, $s_0$ should be identified.

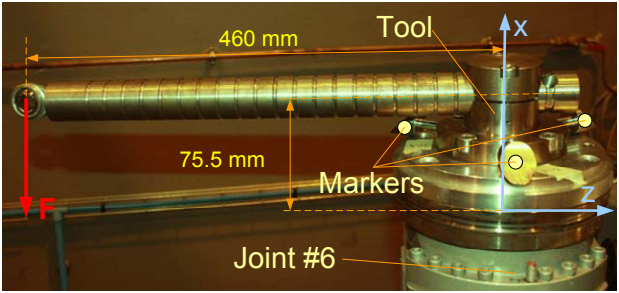

Figure 5. End-effector used for elastostatic calibration experiments

### B. Identification algorithm

To take into account the compensator influence while retaining our previous approach developed for serial robots without compensators, it is proposed to include in the second joint an equivalent virtual spring with non-linear stiffness depending on the joint variable $q_2$ (see eq. (8)). Using this idea, it is convenient to consider several independent parameters $k_{\theta_{2i}}$ corresponding to each value of $q_2$. This allows us to obtain linear form of the identification equations that can be easily solved using standard least-square technique.

Let us denote the set of desired parameters $k_1, (k_{21}, k_{22}...), k_3, ..., k_6$ as the vector $\mathbf{k}$ and linearize (1) with respect to this vector. This allows us to present the relevant force displacement relations in the form

$$\Delta\mathbf{p}_i = \mathbf{B}_i^{(p)} \mathbf{k} \qquad (30)$$

where matrices $\mathbf{B}_i^{(p)}$ are composed of the elements of the matrix

$$\mathbf{A}_i = \left[ \mathbf{J}_{1i} \mathbf{J}_{1i}^T \mathbf{F}_i, ..., \mathbf{J}_{ni} \mathbf{J}_{ni}^T \mathbf{F}_i \right] \quad (i=\overline{1,m}) \qquad (31)$$

that is usually used in stiffness analysis of serial manipulators [14]. Here, $\mathbf{J}_{ni}$ denotes the manipulator Jacobian column, $\mathbf{F}_i$ is the applied external force, and superscript '(p)' stands for the Cartesian coordinates (position without orientation). It is clear that transformation from $\mathbf{A}_i$ to $\mathbf{B}_i^{(p)}$ is rather trivial and is based on the extraction from $\mathbf{A}_i$ the first three lines and inserting in it several zero columns.

Using these notations, the elastostatic parameters identification can be reduced to the following optimization problem

$$F = \sum_{i=1}^{m}(\mathbf{B}_i^{(p)}\mathbf{k} - \Delta\mathbf{p}_i)^T(\mathbf{B}_i^{(p)}\mathbf{k} - \Delta\mathbf{p}_i) \to \min_{k_j, k_c, \rho_0} \qquad (32)$$

which leads to the following solution

$$\mathbf{k} = \left(\sum_{i=1}^{m}\mathbf{B}_i^{(p)^T}\mathbf{B}_i^{(p)}\right)^{-1} \cdot \left(\sum_{i=1}^{m}\mathbf{B}_i^{(p)^T}\Delta\mathbf{p}_i\right) \qquad (33)$$

where the parameters $k_1, k_3, ..., k_6$ describe the compliance of the virtual joints #1,#3,...#6, while the rest of them $k_{21}, k_{22}...$ present an auxiliary dataset allowing to separate the compliance of the joint #2 and the compensator parameters $k_c$, $\rho_0$. Using eq. (8), the desired expressions can be written as

$$\left[ K_{\theta_2}^0 \quad K_c \quad s_0 \cdot K_c \right]^T = \left(\sum_{i=1}^{m_q} \mathbf{C}_i^T \mathbf{C}_i\right)^{-1} \left(\sum_{i=1}^{m_q} \mathbf{C}_i^T K_{\theta_{2i}}\right) \qquad (34)$$

where $m_q$ is the number of different angles $q_2$ in the experimental data,

$$\mathbf{C}_i = \left[ 1 \quad -aL\cos\gamma_i \quad aL/s\left(aL/s^2 \cdot \sin^2\gamma_i + \cos\gamma_i\right) \right] \qquad (35)$$

here $\gamma_i = \alpha - q_{2i}$.

Thus, the proposed modification of the previously developed calibration technique allows us to find the manipulator and compensator parameters simultaneously. An open question, however, is how to find the set of measurement configurations that ensure the lowest impact of the measurement noise.

### C. Design of calibration experiments

The main idea of the calibration experiment design is to select a set of robot configurations $\{\mathbf{q}_i\}$ (and corresponding external loadings $\{\mathbf{F}_i\}$) that ensure the best identification accuracy. The key issue here is the ranging of different plans in accordance with the prescribed performance measure. This problem has been already studied in the classical regression analysis [15], however the results are not suitable for the elastostatic calibration and require additional efforts.

In this work, it is proposed to use the industry oriented performance measure that evaluates the calibration plan quality. Its physical meaning is the robot positioning accuracy (under the loading), which is achieved after compliance error compensation based on the identified elastostatic parameters. It should be noted that usual approach (used in the classical design of experiments) evaluates the quality of experiments via covariance matrix of the identified parameters, which is does not have sense for our application.

Assuming that each experiment includes the additive measurement error $\boldsymbol{\varepsilon}_i$, the covariance matrix for the desired parameters $\mathbf{k}$ can be expressed as

$$\text{cov}(\mathbf{k}) = \left(\sum_{i=1}^{m}\mathbf{B}_i^{(p)^T}\mathbf{B}_i^{(p)}\right)^{-1} \\ \times \text{E}\left(\sum_{i=1}^{m}\mathbf{B}_i^{(p)^T}\boldsymbol{\varepsilon}_i^T\boldsymbol{\varepsilon}_i\mathbf{B}_i^{(p)}\right)\left(\sum_{i=1}^{m}\mathbf{B}_i^{(p)^T}\mathbf{B}_i^{(p)}\right)^{-1} \qquad (36)$$

Following also usual assumption concerning the measurement errors (independent identically distributed, with zero expectation and standard deviation $\sigma^2$ for each coordinate), the above equation can be simplified to

$$\mathrm{cov}(\mathbf{k}) = \sigma^2 \left( \sum_{i=1}^{m} \mathbf{B}_i^{(p)T} \mathbf{B}_i^{(p)} \right)^{-1} \quad (37)$$

Hence, the impact of the measurement errors on the accuracy of the identified parameters $\mathbf{k}$ is defined by the matrix $\sum_{i=1}^{m} \mathbf{B}_i^{(p)T} \mathbf{B}_i^{(p)}$ (in regression analysis it is known as the information matrix).

It is evident that in practice the most essential is not the accuracy of the parameters identification, but the accuracy of the robot positioning achieved using these parameters. Taking into account that this accuracy highly depends on the robot configuration (and varies throughout the workspace), it is proposed to evaluate the calibration accuracy in a certain given "test-pose" provided by the user. For the considered application, the test pose is related to the typical machining configuration $\mathbf{q}_0$ and corresponding external loading $\mathbf{F}_0$ related to the technological process. Let us denote the mean square value of the mentioned positioning error as $\rho_0^2$ and the matrix $\mathbf{A}_i^{(p)}$ (see eq. (31)) corresponding to this test pose as $\mathbf{A}_0^{(p)}$.

It should be noted that that the proposed approach operates with a specific structure of the parameters included in the vector $\mathbf{k}$, where the second joint is presented by several components $k_{21}, k_{22}...$ while the other joints are described by a single parameter $k_1, k_3...k_6$. This motivates further re-arrangement of the vector $\mathbf{k}$ and replacing it by several vectors $\mathbf{k}_j = (k_1, k_{2_j}, k_3, ...k_6)$ of size $6 \times 1$. Using this notation, the above mentioned performance measure can be expressed as

$$\rho_0^2 = \sum_{j=1}^{m_q} \mathrm{E}\left( \delta\mathbf{k}_j^T \mathbf{A}_0^{(p)T} \mathbf{A}_0^{(p)} \delta\mathbf{k}_j \right) \quad (38)$$

where $\delta\mathbf{k}_j$ is the elastostatic parameters estimation error caused by the measurement noise for $q_{2_j}$. Further, after substituting $\delta\mathbf{p}^T \delta\mathbf{p} = \mathrm{trace}(\delta\mathbf{p}\delta\mathbf{p}^T)$ and taking into account that $\mathrm{E}(\delta\mathbf{k}_j\delta\mathbf{k}_j^T) = \mathrm{cov}(\mathbf{k}_j)$, the performance measure $\rho_0^2$ can be presented as

$$\rho_0^2 = \sigma^2 \mathrm{trace}\left( \mathbf{A}_0^{(p)} \sum_{j=1}^{m_q} \left( \sum_{i=1}^{m} \mathbf{A}_i^{j(p)T} \mathbf{A}_i^{j(p)} \right)^{-1} \mathbf{A}_0^{(p)T} \right) \quad (39)$$

Based on this performance measure, the calibration experiment design can be reduced to the following optimization problem

$$\mathrm{trace}\left( \mathbf{A}_0^{(p)} \sum_{j=1}^{m_q} \left( \sum_{i=1}^{m} \mathbf{A}_i^{j(p)T} \mathbf{A}_i^{j(p)} \right)^{-1} \mathbf{A}_0^{(p)T} \right) \to \min_{\{\mathbf{q}_i, \mathbf{F}_i\}} \quad (40)$$

subject to

$$\|\mathbf{F}_i\| < F_{\max}, \quad i = 1..m \quad (41)$$

whose solution gives a set of the desired manipulator configurations and corresponding external loadings. It is evident that its analytical solution can hardly be obtained and a numerical approach is the only reasonable one. More detailed description of the developed technique providing generation of the optimal calibration plans is presented in our previous publication [14], where the problem of gravity compensator modeling had not been addressed yet.

### D. Experimental results

Similar to the previous section, the developed technique has been applied to the robot KR-270. The parameters of interest were the compliances $k_j$ of the actuated joints and the elastostatic parameters $k_c$, $s_0$ of the gravity compensator. To generate elastostatic deflections, the gravity forces 250-280 kg have been applied to the robot end-effector (see Fig 6). The Cartesian coordinates of three markers located on the tool (see Fig, 5) have been measured before and after the loading.

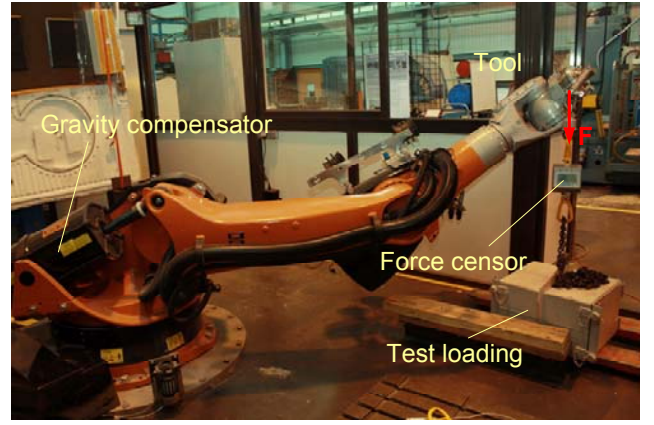

Figure 6. Experimental setup for the identification of the elastostatic parameters

To find optimal measurement configurations, the design of experiments has been carried out for five different angles $q_2$ that are distributed between the joint limits. For each $q_2$ three optimal measurement configurations have been found taking into account physical constraints that are related to the joint limits and the possibility to apply the gravity force (work-cell obstacles and safety reasons). The results of the calibration experiment design are presented in Table III. Here, values of $q_1$ have been chosen to ensure good visibility of the markers for the laser tracker. To ensure identification accuracy for each configuration, the experiments were repeated three times. In total, 405 equations were considered for the identification, from which 7 physical parameters have been obtained (because of page limit, the experimental data cannot be presented in the paper).

TABLE III. OPTIMAL MEASUREMENT CONFIGURATIONS

| Joint angles, [deg] | | | | | |
|---|---|---|---|---|---|
| $q_1$ | $q_2$ | $q_3$ | $q_4$ | $q_5$ | $q_6$ |
| 79.20 |  | -5.57 | 51.00 | -97.52 | -91.67 |
| 63.00 | -0.01 | -12.22 | -56.49 | 41.42 | 150.55 |
| 63.00 |  | -47.98 | -70.04 | -61.55 | 177.16 |
| 95.00 |  | 33.00 | 129.69 | -98.10 | 90.57 |
| 95.00 | -25.24 | -107.01 | 109.95 | -61.19 | 174.21 |
| 105.00 |  | 14.30 | 55.21 | 41.26 | -152.97 |
| 56.60 |  | 44.54 | -55.11 | 41.90 | 152.06 |
| 56.60 | -56.9 | 64.73 | -129.65 | -98.260 | -90.55 |
| 144.80 |  | 104.49 | -69.41 | 61.67 | -6.33 |
| -41.00 |  | -91.68 | 55.12 | 41.53 | -152.48 |
| -143.00 | -99.85 | -32.64 | 110.31 | -61.47 | -6.29 |
| -143.00 |  | -72.01 | 129.65 | -98.09 | 90.82 |
| 133.00 |  | 147.68 | 129.64 | -97.90 | 90.99 |
| -60.00 | -140 | 7.59 | -110.09 | -61.36 | -174.09 |
| -60.00 |  | -52.00 | -124.89 | -41.62 | 27.78 |

TABLE IV. ELASTO-STATIC PARAMETERS OF ROBOT KUKA KR-270

| Parameter | value | CI |
|---|---|---|
| $k_c$, [rad×μm/N] | 0.144 | ±0.031 (21.5%) |
| $s_0$, [mm] | 458 | ±27 (5.9%) |
| $k_2$, [rad×μm/N] | 0.302 | ±0.004 (1.3%) |
| $k_3$, [rad×μm/N] | 0.406 | ±0.008 (2.0%) |
| $k_4$, [rad×μm/N] | 3.002 | ±0.115 (3.8%) |
| $k_5$, [rad×μm/N] | 3.303 | ±0.162 (4.9%) |
| $k_6$, [rad×μm/N] | 2.365 | ±0.095 (4.0%) |

The obtained experimental data have been processed using the identification algorithm presented in sub-section V.B. Corresponding values of the gravity compensator and the manipulator elastostatic parameters are presented in Table IV. It also includes the confidence intervals computed as $\pm 3\sigma$, where the standard deviation $\sigma$ has been evaluated using Gibbs sampling. Using the obtained results it is possible to identify an equivalent non-linear spring $k_2(q_2)$ (see Fig, 7) that is used in the stiffness modeling of the manipulator with the gravity compensator (see Section II).

The identified joint compliances can be used to predict robot deformations under the external loading. The identification errors of the joint compliances vary from 1.3% to 4.9%, where the lowest errors have been achieved for the joints #2 and #3. The highest errors are in the joints #4-#6. So, higher precision has been achieved for the joints that are closer located to the manipulator base. This fact is due to the different joint errors contribution to the total positioning error, which have been minimized while design of calibration experiments. Comparatively low identification accuracy for the compensator spring is caused by limited number of different angles $q_2$. Fig. 7 shows that due to the gravity compensator the equivalent compliance of the second joint has been decreased comparing to the compliance of the second joint of the corresponding serial manipulator.

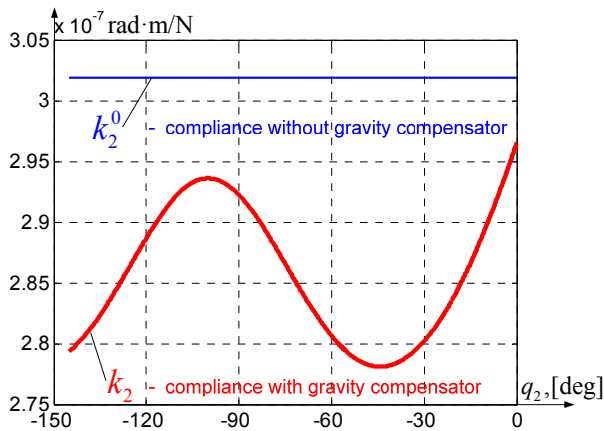

Figure 7. Compliance of equivalent non-linear spring in the second joint

## VI. CONCLUSIONS

The paper presents a new approach for the identification of the elastostatic parameters of heavy industrial robots with the gravity compensator. It proposes a methodology and data processing algorithms for the identification of both geometrical and elastostatic parameters of gravity compensator and manipulator. To increase the identification accuracy, the design of experiments has been used aimed at proper selection of the measurement configurations. In contrast to other works, it is based on a new industry oriented performance measure that is related to the robot accuracy under the loading.

The advantages of the developed techniques are illustrated by experimental study of the industrial robot Kuka KR-270, for which the joint compliances and parameters of the gravity compensator have been identified.

ACKNOWLEDGMENT

The work presented in this paper was partially funded by the ANR, France (Project ANR-2010-SEGI-003-02-COROUSSO). The authors also thank Fabien Truchet, Guillaume Gallot, Joachim Marais and Sébastien Garnier for their great help with the experiments.

REFERENCES

[1] M.A. Meggiolaro, S. Dubowsky, C. Mavroidis, "Geometric and elastic error calibration of a high accuracy patient positioning system," Mechanism and Machine Theory, vol. 40, 415–427, 2005.
[2] J. Kövecses, J. Angeles, "The stiffness matrix in elastically articulated rigid-body systems," Multibody System Dynamics (2007) pp. 169–184, 2007.
[3] B.-J. Yi, R.A. Freeman, "Geometric analysis antagonistic stiffness redundantly actuated parallel mechanism", Journal of Robotic Systems vol. 10(5), pp. 581-603, 1993
[4] J. Salisbury, "Active stiffness control of a manipulator in Cartesian coordinates," in Proc. 19th IEEE Conf. Decision Control, 1980, pp. 87–97.
[5] N. Takesue, T. Ikematsu, H. Murayama and H. Fujimoto, "Design and Prototype of Variable Gravity Compensation Mechanism (VGCM)," Journal of Robotics and Mechatronics, Vol. 23, No. 2, pp. 249-257, 2011.
[6] A. De Luca, F. Flacco, "A PD-type Regulator with Exact Gravity Cancellation for Robots with Flexible Joints," in Proc. 2011 IEEE International Conference on Robotics and Automation, pp. 317-323
[7] O. Company, F. Pierrot, J.-C. Fauroux, "A Method for Modeling Analytical Stiffness of a Lower Mobility Parallel Manipulator," in: Proceedings of IEEE International Conference on Robotics and Automation (ICRA), 2005, pp. 3232 - 3237
[8] J.-P. Merlet, C. Gosselin, Parallel mechanisms and robots, In B. Siciliano, O. Khatib, (Eds.), Handbook of robotics, Springer, Berlin, 2008, pp. 269-285.
[9] B.C. Bouzgarrou, J.C. Fauroux, G. Gogu, and Y. Heerah, "Rigidity analysis of T3R1 parallel robot with uncoupled kinematics," Proc. of the 35th International Symposium on Robotics, Paris, France, 2004.
[10] A. Pashkevich, A. Klimchik, D. Chablat, "Enhanced stiffness modeling of manipulators with passive joints," Mechanism and Machine Theory 46(2011) 662-679.
[11] A. Klimchik, D. Bondarenko, A. Pashkevich, S. Briot, B. Furet, "Compensation of tool deflection in robotic-based Milling," the 9th International Conference on Informatics in Control, Automation and Robotics (ICINCO 2012), July 28-31, 2012, Rome, Italy, pp. 113-122.
[12] D. Daney, N. Andreff, G. Chabert, Y. Papegay, "Interval method for calibration of parallel robots: Vision-based experiments," Mechanism and Machine Theory, vol. 41, 2006, pp. 929-944.
[13] Hollerbach J., Khalil W., Gautier M., Springer Handbook of robotics, Springer, 2008, "Chapter: Model identification," pp. 321-344.
[14] A. Klimchik, Y. Wu, A. Pashkevich, S. Caro, B. Furet, "Optimal Selection of Measurement Configurations for Stiffness Model Calibration of Anthropomorphic Manipulators," Applied Mechanics and Materials, Volume 162, 2012, pp. 161-170
[15] Atkinson A., Done A., Optimum Experi-ment Designs. Oxford University Press, 1992